\documentclass{CVM}

\usepackage{amsmath}
\usepackage{color}
\usepackage{multirow}
\usepackage{colortbl}
\usepackage{graphbox}
\newcommand{\FLIP}{\protect\reflectbox{F}LIP\xspace}
\definecolor{mygray}{gray}{.97}

\CVMsetup{
type      = {Research Article},
doi       = {s41095-0xx-xxxx-x},
title     = {A Simple And Effective Filtering Scheme For Improving Neural Fields},
author    = {Yixin Zhuang$^{1}$\cor{}\\
},
runauthor = {Yixin Zhuang},
abstract  = {
Recently, neural fields, also known as coordinate-based MLPs, have achieved impressive results in representing low-dimensional data. Unlike CNN, MLPs are globally connected and lack local control; adjusting a local region leads to global changes. Therefore, improving local neural fields usually leads to a dilemma: filtering out local artifacts can simultaneously smooth away desired details. Our solution is a new filtering technique that consists of two counteractive operators: a smoothing operator that provides global smoothing for better generalization, and conversely a recovering operator that provides better controllability for local adjustments. We have found that using either operator alone can lead to an increase in noisy artifacts or oversmoothed regions. By combining the two operators, \emph{smoothing} and \emph{sharpening} can be adjusted to first smooth the entire region and then recover fine-grained details in regions overly smoothed. In this way, our filter helps neural fields remove much noise while enhancing details. We demonstrate the benefits of our filter on various tasks and show significant improvements over state-of-the-art methods. Moreover, our filter also provides better performance in terms of convergence speed and network stability. Code is avaialble at (\url{https://github.com/yixin26/FINN}).

},
keywords  = {Neural Fields, Neural Filter, Implicit Neural Representation, Representation Learning},
copyright = {The Author(s)},
}

\begin{document}

\maketitle

\section{Introduction}

\begin{figure}[t!]
\begin{center}
{\includegraphics[width=1.0\linewidth]{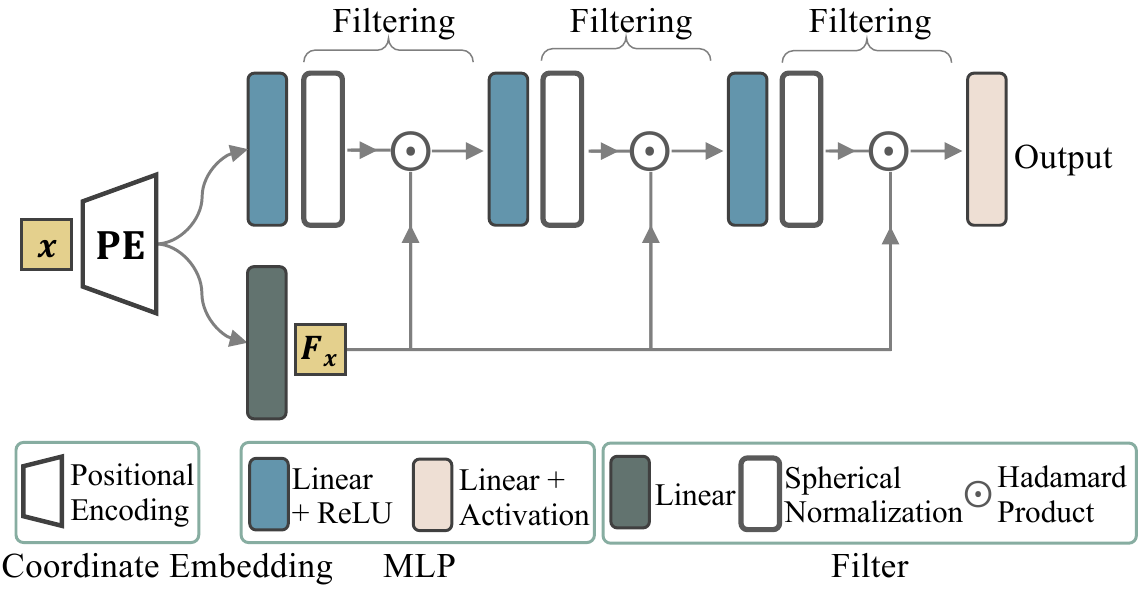} }
\end{center}
\caption{
We introduce a new filter for coordinate-based MLPs. It consists of two operators, a \emph{smoothing} operator, i.e., a spherical normalization that provides global smoothing, and a \emph{recovering} operator, i.e., a linear transformation that controls local sharpening. Promoting MLPs with either operator alone can lead to oversmoothing or overfitting. The key idea is to combine the two operators so that MLPs can adjust between \emph{smoothing} and \emph{sharpening} to remove noisy artifacts while enhancing details.
}
\label{fig:network}
\end{figure}

Neural fields are emerging as powerful representations for visual content~\cite{NeuralFields}.
They are implicit functions learned via coordinate-based MLPs that map spatial coordinates to their corresponding values, e.g., RGB and signed distance.
Due to the spectral bias of neural networks~\cite{basri20a,rahaman19a}, ReLU-MLPs tend to learn the low-frequency components of signals and require more network capacity and training time to adapt the high-frequency components.
Learning high frequencies can be facilitated by using Fourier features to embed input coordinates (FFN)~\cite{tancik2020fourfeat} or by replacing ReLU activation functions with periodic functions in MLPs (SIREN)~\cite{sitzmann2019siren}.
They shift the spectral bias with frequencies that better match those of the input data.

\begin{figure*}[t!]
\begin{center}
{\includegraphics[width=1.0\linewidth]{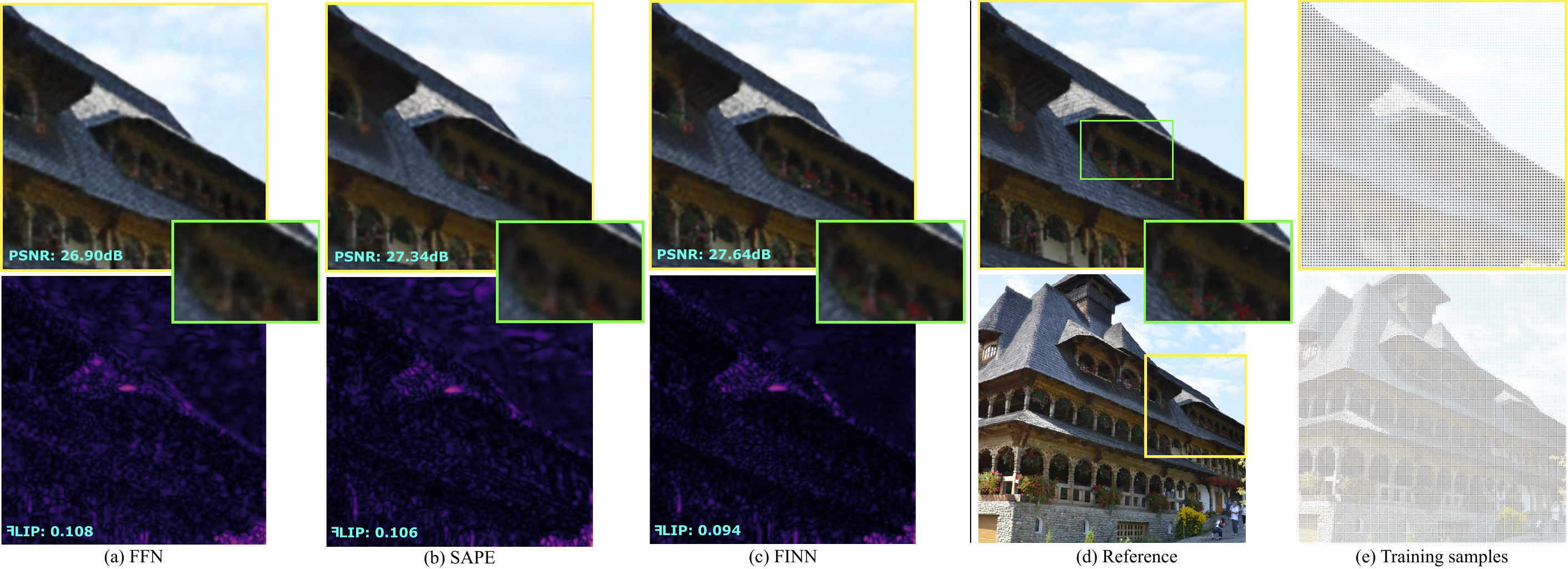} }
\end{center}
\caption{
(a) An image generated with FFN contains random noise, which is highlighted in the error map from FLIP (bottom). (b) SAPE smooths out much of the random noise by preventing the use of excessively high local frequency parameters at any spatial position, but at the cost of poorly fitting small structures such as clouds and roofs. 
(c) Our method incorporates a filter in FFN that can effectively remove much noise while better fitting details, i.e., having fewer errors in both flat and sharp regions (which is better revealed by the heat maps). 
We highlight some areas of the images in yellow and green boxes and show PSNR and \FLIP at the bottom of each image. The input (e) is an image with a resolution of 256$\times$256 sampled regularly from (d) and the outputs are images with a resolution of 512$\times$512. 
}
\label{fig:comp_teaser}
\end{figure*}

While very effective at fitting complicated signals, FFN and SIREN may use frequencies that are too high, transforming the entire space from linear/smooth to highly nonlinear.
This leads to unexpected random variations in the unseen space between training samples and affects generalization.
To ensure global smoothing, SAPE~\cite{hertz2021sape} proposed to use local frequency parameters for individual spatial locations and learn to avoid excessively high variations around the training samples.
Meanwhile, ModSIREN~\cite{Mehta21} modulates the frequency parameters based on the spatial grids.
MFN~\cite{FathonySWK21} and BACON~\cite{lindell2021bacon} and pi-GAN~\cite{chanmonteiro2020pi-GAN} also adjusts the local frequency and possibly the phase and amplification parameters of periodic functions to better fit individual data.
An example is shown in Figure~\ref{fig:comp_teaser}, where the local frequency adjustment method (i.e., SAPE) produces results with less random noise compared to the baseline method (i.e., FFN).

Adjusting either global or local frequency parameters in MLPs can lead to large global changes in outputs because MLPs are globally connected and each neuron affects all neurons in subsequent layers. Therefore, it is not possible to change one local area without affecting others, which inevitably leads to a dilemma: filtering out local artifacts can simultaneously smooth away desired details.
For example, SAPE tends to oversmooth the result (e.g., in Figure~\ref{fig:comp_teaser} the details in the sky and on the roofs are lost), even though it reduces a lot of noise.

Similar to how CNN use filters to control the smoothness and sharpness of local regions, we investigate a new filtering scheme for MLPs with similar controllability.
We study two types of filtering functions: the smoothing operator, which reduces variations, and the recovering operator, which increases variations.
We develop the smoothing operator as spherical normalization\footnote{
Spherical normalization is the normalization of inputs to the hypersphere. We define this term to distinguish it from other normalization methods in this context.}
that provides global smoothing, and the recovering operator as a linear transformation for better local fitting.
The operators can be used in conventional MLPs or in the aforementioned works that can manage frequencies. Figure~\ref{fig:network} shows the network using our filter in FFN.

\begin{figure}[t!]
\begin{center}
{\includegraphics[width=1.0\linewidth]{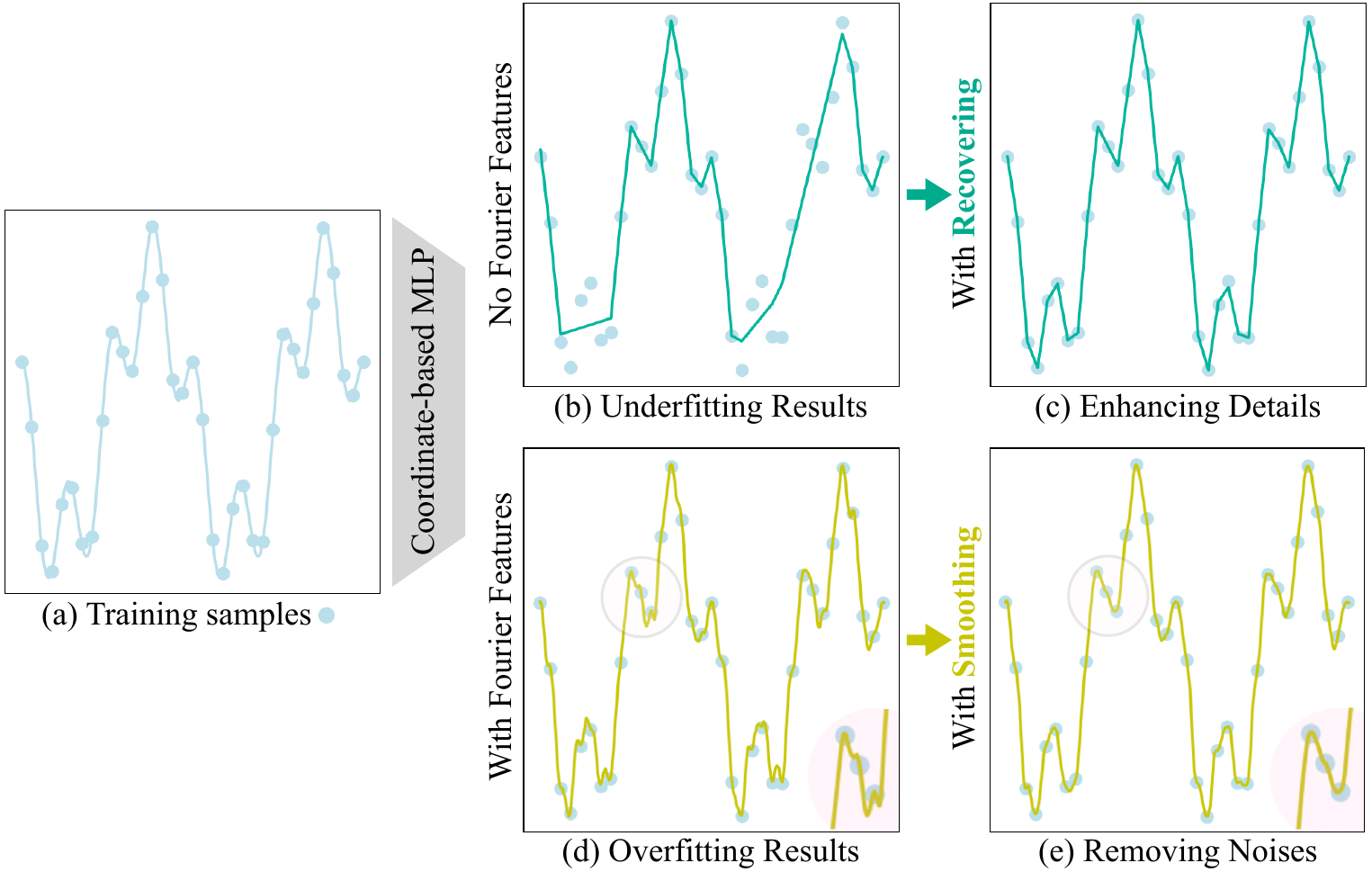} }
\end{center}
\caption{
A 1D toy example to illustrate the effects of smoothing and recovering operators. When reconstructing a 1D signal (a) with coordinate-based MLPs, underfitting (b) may occur if the capacity of the MLPs is insufficient, or overfitting (d) may occur if too high a frequency parameter is used in the FFN. Without increasing network parameters or adjusting frequency parameters, the proposed recovering operator helps to recover structures (c), and the smoothing operator removes noisy artifacts (e).
}
\label{fig:teaser}
\end{figure}

The filtering effect of each operator is shown in Figure~\ref{fig:teaser}. For a 1D example in (a), a small ReLU-MLP in (b) gives inadequate results, which can be greatly improved by a recovering operator in (c). Note that (b)\&(c) have similar network capacity. When fitting the same samples with FFN (d), the result exhibits unexpected bumps (local high frequencies) between nearby samples, which can be effectively removed by the smoothing operator (e). Note that (d)\&(e) use the same global frequency parameter.

For more complicated signals, such as images and shapes, neither operator alone can effectively remove noisy artifacts and recover fine details.
The smoothing operator ensures that the entire space is smoothed, and is therefore unable to enhance fine content. 
Theoretically, the recovering has more control over the variations of the coordinates, which can perform smoothing and sharpening by reducing or increasing the scales in the linear transformation.
In practice, however, it only has a sharpening effect to better fit the training samples, and for the unseen space it has no intention to reduce the variations and leave the random noise there.
{Therefore, without a smoothing constraint, the linear transformation alone cannot eliminate the noise in the unseen space.}
For example, MFN and BACON also use a multiplicative operation (a linear transformation) that scales local variations. This gives them a strong performance improvement in training, but little improvement in generalization compared to FFN and SIREN.

To adjust between smoothing and sharpening, we need both operators together, not just one. By first smoothing the entire region and then recovering the fine details in the oversmoothed regions, the two coounteractive operators work together to achieve a better fit to the training samples and also smooth the unseen space, leading to better fitting and generalization.
Figure~\ref{fig:comp_teaser} shows the filtering effect of our method, which has less noise and finer details than the methods without filter.

Our smoothing and recovering operators must be carefully balanced. Since the former is parameter-free and unchanged, the latter must be designed so that it is not too strong, which makes the smoothing operator less effective, or too weak, which cannot sufficiently improve the details. We will explore different designs of the recovering operator to coordinate with the smoothing operator.

To demonstrate the effectiveness of our filter, we designed networks using the filter for different tasks. Experimental results show that our method performs better than the state-of-the-art methods.
Moreover, our filter design leads to faster convergence speed and better network stability through SGD optimization.
Our contributions can be summarized as follows.
\begin{itemize}
\item A new filtering scheme that combines two counteractive operators to adjust smoothing and sharpening in MLPs. The filter helps neural fields remove noisy artifacts while enhancing distinct details.
\item Better performance over state-of-the-art methods on multiple tasks, including image representation, 3D shape reconstruction and novel view synthesis.
\end{itemize}

\begin{figure*}[tp!]
\begin{center}
{\includegraphics[width=1.0\linewidth]{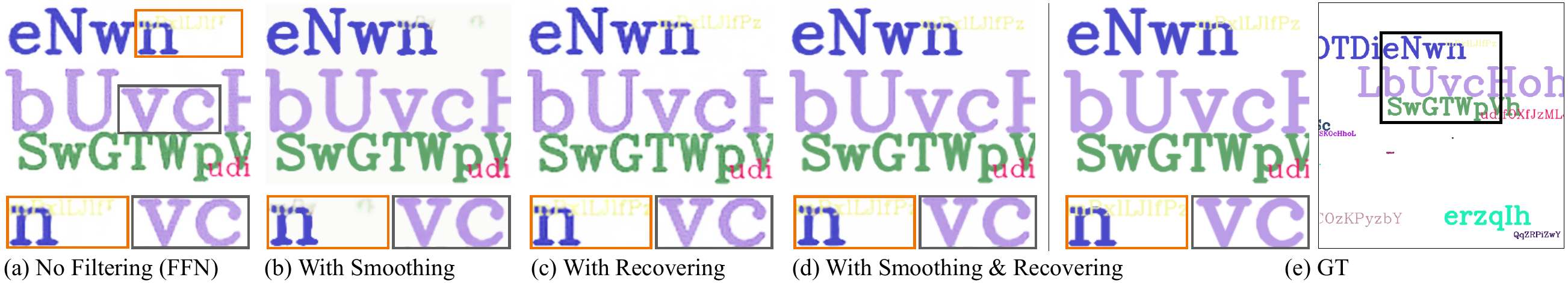} }
\end{center}
\caption{
Illustration of the filtering effect on a 2D image. The target image (e) contains homogeneous regions everywhere and some small-scale text. We show only a subset of (e) to illustrate the difference between the methods. (a) is generated by FFN without filters and shows random noise (highlighted in gray box). Smoothing (b) provides smoother text by removing the noise in it, but this oversmooths some small-scale structures (as shown in the orange box). Recovering (c) can restore finer details for small text, but is not able to smooth the noise. The combination of two operators (d) allows smoothing noise and restoring small structures.
}
\label{fig:idea2}
\end{figure*}

\section{Related Work}

\paragraph{Implicit Neural Representation.}
Deep neural networks have been shown to be effective for learning implicit functions representing images~\cite{tancik2020fourfeat,sitzmann2019siren,MildenhallSTBRN20,chen2021learning}, {fonts~\cite{reddy2021multi} and 3D humans, objects and scenes~\cite{park2019deepsdf,OccNet,chen2018implicit_decoder,AtzmonL20,GroppYHAL20,SitzmannZW19,peng2021neural}.}
They use coordinate-based MLPs that can be sampled at arbitrarily high spatial resolution. Therefore, such a representation can be used directly for super-resolution tasks. Other applications include view synthesis~\cite{MildenhallSTBRN20,Martin-BruallaR21,yu2020pixelnerf,Bemana2020xfields}, point-cloud-based 3D reconstruction~\cite{AtzmonL20,GroppYHAL20,WilliamsTBZ21,WilliamsSSZBP19}, and 3D reconstruction from single images~\cite{OccNet,DISN,XuFYS20,saito2019pifu}. In addition to visual reconstruction and generation, implicit representation is also widely used for many other tasks, such as feature matching~\cite{ChenFBMG21} and scene understanding~\cite{ZhiICCV2021}. A comprehensive review of the use of implicit representations has been provided by~\cite{NeuralFields}.

As evidenced by~\cite{basri20a,rahaman19a},
ReLU-MLPs have difficulty capturing very detailed signals due to the spectral bias. FFN~\cite{tancik2020fourfeat} uses positional encoding to map input coordinates of signals to Fourier features using sinusoidal functions. 
In SIREN~\cite{sitzmann2019siren}, on the other hand, the ReLU activations in the MLPs are replaced by sine functions. They have a similar spirit that the input or intermediate results are manipulated in the frequency domain to capture high frequencies in the output. Moreover, the sine functions in Fourier features or MLPs  are designed to be learned to better fit individual data~\cite{FathonySWK21,Mehta21}.
In contrast to the use of sine functions, spline positional encoding (SPE) ~\cite{WangLYT21} explores learnable spline functions for coordinate embedding. With sufficient local support of splines, SPE can also approximate the signal with high frequencies. However, when using a small number of local supports, the boundaries become noticeable, resulting in strip noise and significantly reducing visual quality.

Reconstruction of high-frequency details is usually accompanied by the appearance of visual artifacts in the results. Some recent developments have led to structured or hierarchical designs that can further close the gap between the generated results and the target function. They divide the complex functions of 3D objects and scenes~\cite{ChabraLISSLN20,GenovaCSSF20,JiangSMHNF20} or images~\cite{Mehta21} into regular sub-regions and fitting each sub-region while maintaining global consistency.

Subdividing regions may have little effect if a subregion is itself complex. Recently, spatial adaptation on frequencies seems to be a better solution. SAPE~\cite{hertz2021sape} presents a progressive optimization strategy to encode signals with increasing frequencies at single spatial locations. The method reduces noisy artifacts, but tends to produce oversmoothed regions.
Some existing methods such as MFN~\cite{FathonySWK21} and BACON~\cite{lindell2021bacon}, pi-GAN~\cite{chanmonteiro2020pi-GAN}, and ModSIREN~\cite{Mehta21} also modulate parameters of sinusoidal functions for better reconstruction and conditional generation.
As discussed earlier, adjusting these parameters can lead to global and potentially large changes in the results, so they lack local controls for adjustment. Instead of modulating these hyperparameters, we adopt the idea of CNN with filters to provide MLPs with more local controllability.

\paragraph{Deep Image Filters.}
Image filters usually compute the weighted average of the neighboring pixels of the image as output or sometimes use regularization constraints for image optimization.
Recently, a number of researchers have introduced neural network filters that can be learned from a large number of datasets for various applications~\cite{JampaniKG16,ZhangPLL017,YoonJYLK15,harbiCBHD17,ChenXK17,LiHAY19}, such as image denoising and deblurring, etc. They are mainly based on convolutional neural networks, from which the local neighborhood of the query point is determined.
For continuous functions, it is difficult to determine the neighborhood unless one discretizes the input domain with predefined scales or resolutions.
Therefore, we attempt to develop a general filtering scheme for MLPs that can handle image functions and more continuous signals.

\section{Method}
\label{sec:method}

In this section, we introduce neural fields and our filter in section~\ref{sec:imnet} and section~\ref{sec:filtering}, respectively.

\subsection{Neural Implicit Functions}
\label{sec:imnet}

An implicit function is a continuous function $f_{\theta}:\mathbb{R}^a \rightarrow \mathbb{R}^n$ that takes as input a coordinate of any query point from the Euclidean space $x \in \mathbb{R}^a$ and predicts a value in the target space $\mathbb{R}^n$. 
Learning $f_{\theta}$ with a neural network requires a set of coordinate samples as input and corresponding values as output. 
Examples of $f_{\theta}$ include mapping of pixels to RGB values for image functions or the projection of 3D coordinates to signed distance values for 3D shape functions.

The implicit function $f_{\theta}$ is usually modeled with MLPs. Due to the spectral bias, ReLU-based MLPs are difficult to fit high-frequency signals, resulting in severe underfitting. To shift the preference to the higher frequency spectrum, FFN embeds the input in the frequency domain so that MLPs can easily learn high frequencies. Specifically, the coordinate $x$ is represented by a \textit{d}-dimensional Fourier feature vector, $\gamma(x) \in \mathbb{R}^d$, as follows:
\begin{equation}
\gamma(x) = \frac{s}{\sqrt{d}}[\text{cos}(2\pi x B^T) \| \text{sin}(2\pi x B^T) ]
\label{equ1}
\end{equation}
where $\|$ is the concatenation of two vectors and $B$ is a $\frac{d}{2} \times a$ matrix drawn randomly from the Gaussian distribution with standard deviation $\sigma$. $\sigma$ is the key parameter controlling the global frequency and complexity of the results. $s$ is a constant for global scaling.

Using Fourier features facilitates learning of complex signals, but runs the risk that the reconstructed function will have unexpected, dramatic local variations that create random noise.
Noisy artifacts can be removed by reducing the frequency parameter $\sigma$, but this results in oversmoothing of the entire domain. In addition to tuning $\sigma$, we introduce a  filter that can remove noise while improving detail.

\begin{figure*}[tp!]
\begin{center}
{\includegraphics[width=1.0\linewidth]{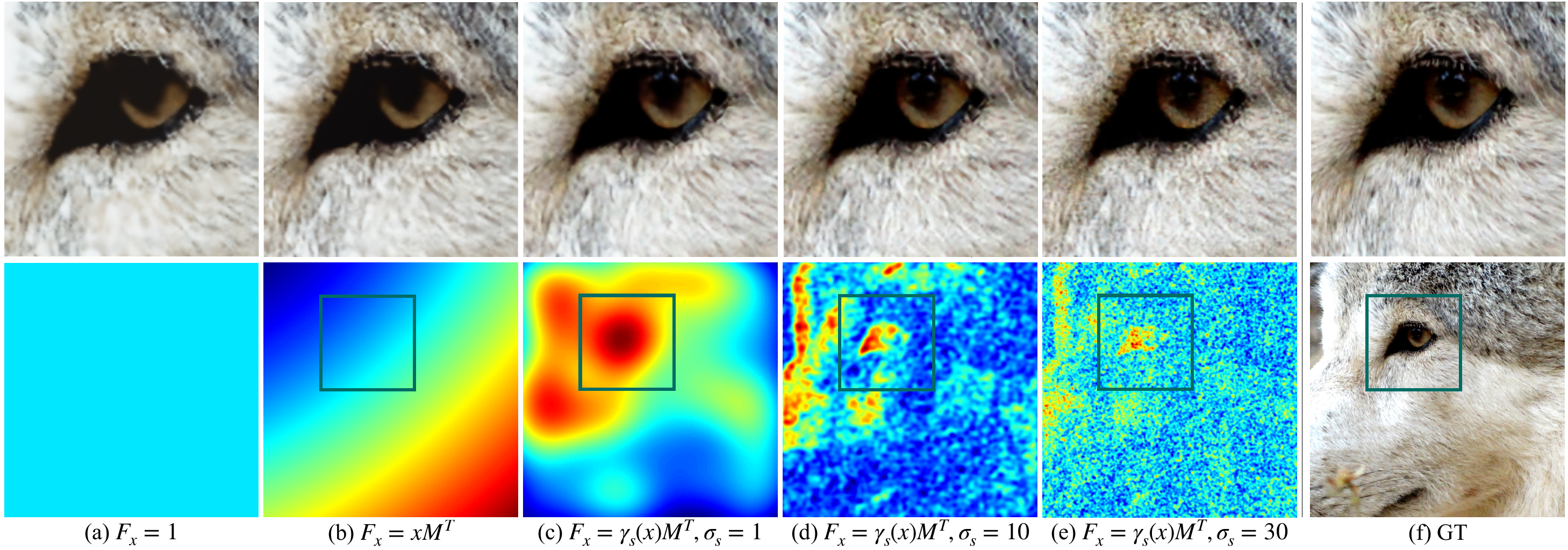} }
\end{center}
\caption{
Illustration of the recovering strength. We show the recovering strength (bottom), i.e., the scaling vectors $F_x$, and the corresponding reconstructed results (top). The reconstruction results show different levels of sharpness depending on $F_x$, which can be a constant (a), or generated from the coordinates $x$ (b), or derived from Fourier features $\gamma_s(x)$ (c-e). 
(a)\&(b) have a problem with oversmoothing, while (c-e) can provide more detail.
In (c-e), using a larger $\sigma_s$ adds more variation to the scaling vectors, resulting in sharper reconstruction, but too high $\sigma_s$ produces more random noise. (d) has a balanced recovering strength that achieves the best PSNR.
}
\label{fig:filteringeffect}
\end{figure*}

\subsection{Filtering Functions}
\label{sec:filtering}

Let $\theta_i, i=1,2,...,k$ denotes the layers of a $k$-layer MLP, then the coordinate $x$ passing through the MLPs produces a sequence of outputs, denoted $y_i, i=1,2,...,k$. We apply a filter to the intermediate outputs of the MLPs, i.e., $y_i, i=1,2,...,k-1$, except for the final output $y_k$.
The network function can be written in recursion as follows:
\begin{equation}
\begin{aligned}
y_1 &= \theta_1(\gamma(x)) \\
y_i &= o(\theta_i(y_{i-1}))\odot F_x, i=2,...,k-1\\
f_{\theta} &= \theta_k(y_{k-1})
\end{aligned}
\label{equ2}
\end{equation}
where $o(\cdot)$ is the spherical normalization function such that $o(v)=\frac{v}{\|v\|}$, and $\odot$ is the Hadamard product multiplying the normalized $y_i$ by a scaling vector $F_x$.
$o(\cdot)$ and $\odot$ denote our smoothing and recovering operators, respectively, and the scaling vector $F_x$ is used to control the sharpness of the recovering at each location $x$. We define $F_x$ as
\begin{equation}
\begin{aligned}
F_x &= \gamma_s(x)M^T
\end{aligned}
\label{equ3}
\end{equation}
where $F_x$ is generated from another Fourier feature $\gamma_s(x)$ using a linear transformation $M$. We denote the control parameter for $\gamma_s(x)$ as $\sigma_s$.
Note that $\gamma_s(x)$ serves for $F_x$ and is different from $\gamma(x)$ for MLPs, but usually they can be identical.
If they are different, i.e., $\gamma_s(x)$ and $\gamma(x)$ use different frequency parameters, then MLPs are able to adjust local variations over a larger range, between $\sigma_s$ and $\sigma$. In practice, we set $\gamma_s(x)$ and $\gamma(x)$ to identical for the representation of image and shape functions, and to different for more complex functions, such as the neural radiance field that mixes color and density fields.

A 2D example in Figure~\ref{fig:idea2} illustrates the effects of the smoothing and recovering operators and their combination. The baseline method (a) has no filter and produces random noise and oversmoothed patches. Smoothing (b) removes most of the noise, but loses some small-scale text. Recovering (c) reveals many small structures, but cannot reduce noise. Combining both operators (d) can restore fine-grained details and remove noise at the same time.

\paragraph{{Smoothing Constraint.}}
{There are several ways to smooth the output of neural networks, by using a smoothing operator that computes the weighted average of the neighboring samples as output, or by applying a smoothing regularization in the loss function. The latter, the smoothing constraints, can be defined for a single sample, e.g., by applying the Eikonal equation to individual samples, normalizing the magnitude of each sample's gradient to 1. Similarly, our smoothing operator can be understood as a smoothing constraint applied to intermediate features of MLPs, forcing the magnitude of the features to a constant such that the resulting feature space is a hypersphere that is smooth everywhere. Once all intermediate MLP features are smoothed, the final output is also smoothed everywhere by a certain degree, resulting in a global smoothing effect of the results. This operator is therefore referred to as the smoothing operator in our method, and serves as a smoothing constraint on the neural fields.
}

\paragraph{Recovering Strength.}
One can change the scaling vector $F_x$ to control the details brought by the recovering operator. For example, smoother reconstruction requires weaker recovering in which $F_x$ is generated from a flatter domain. 
In Figure~\ref{fig:filteringeffect}, several $F_x$ are shown in the bottom row, including $F_x=1, xM^T$ and $\gamma_s(x)M^T$ ($\sigma_s=1,10$ and $30$) with increasing complexity, and the corresponding results reveal different degrees of sharpness. $F_x=1$ means that recovering operator is not used, and a too strong recovering operator ($\sigma_s=30$) cancels the smoothing effect.
The best results in this example are obtained when $\gamma_s(x)$ and $\gamma(x)$ are identical, i.e., $\sigma_s=\sigma=10$. 
In most of our experiments we use the default setting $F_x=\gamma_s(x)M^T$, $\sigma_s=\sigma$.

\begin{figure*}[h]
\begin{center}
{\includegraphics[width=0.99\linewidth]{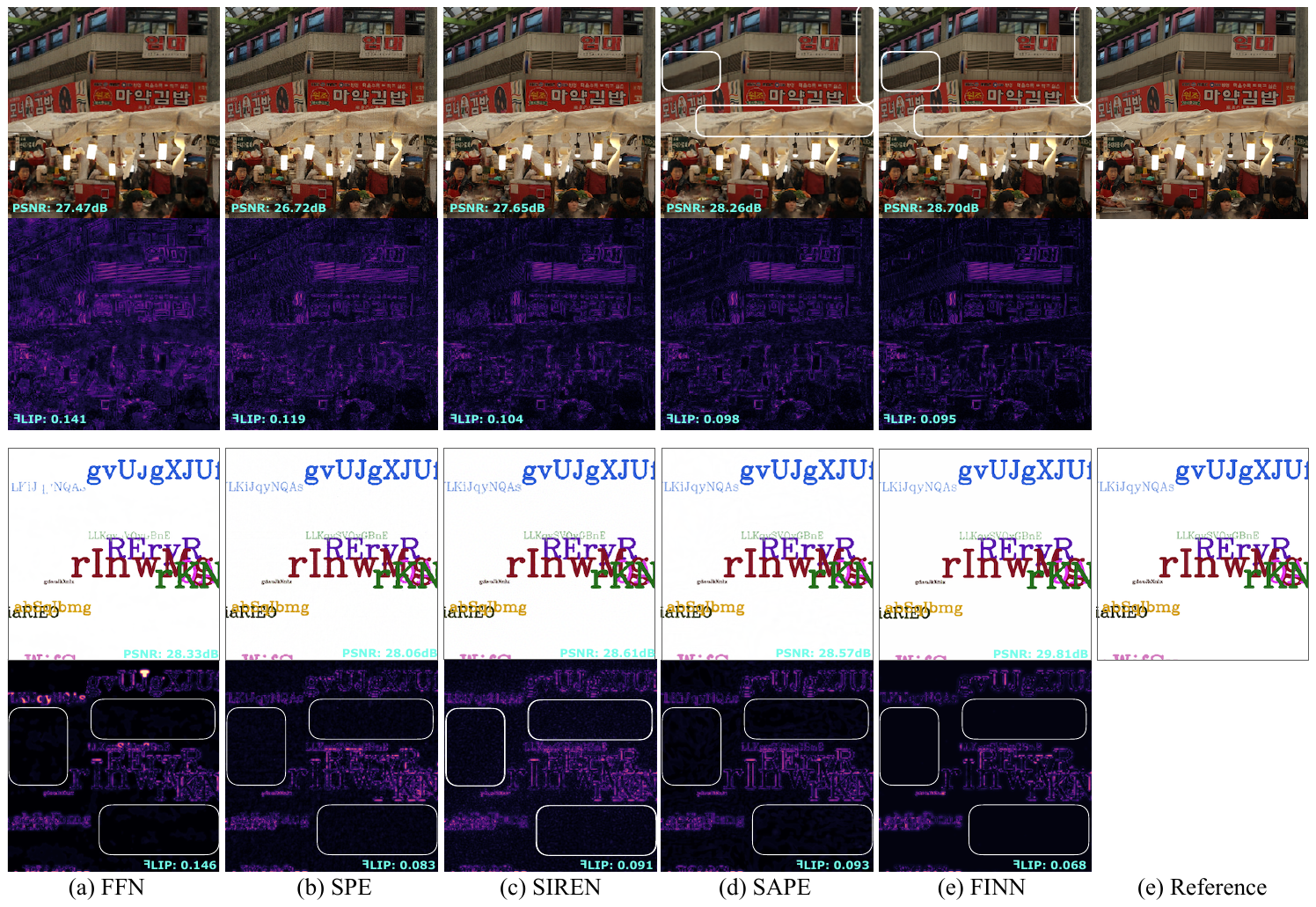} }
\end{center}
\caption{
Qualitative image generalization results. Random noise is evident in FFN, SPE, and SIREN in both ``natural" (top) and ``text" (bottom) images. SAPE removes much of the noise, but oversmooths some areas (e.g., in the highlighted boxes). Using the error map, we can also see a lot of weak noise in the background of the ``text" image in FFN, SPE, SIREN, and SAPE.}\label{fig:2dresults}
\end{figure*}

\section{Experiments} 
\label{sec:exp}

We develop networks (denoted as FINN) using our filter for 2D image representation, 3D shape reconstruction, and novel view synthesis, and validate the benefits of the filter.

\subsection{Image Representation}
Image functions map a 2D coordinate to an RGB color. Our network has 512-dimensional Fourier features (with $\sigma=10, s=80$ for $\gamma(x)$ and $\gamma_s(x)$ is identical to $\gamma(x)$), MLPs with 3 hidden layers, 256 channels, ReLU activation, and sigmoid at the output. In addition, our filter contains a 512$\times$256 matrix to generate the scaling vector $F_x$. We use MSE loss to train the network for 2000 epochs with a learning rate of 1e-3.

We compare our method with FFN, SIREN, SPE, and SAPE, on the datasets of ``natural" and ``text" images from FFN.
The training pixels are sampled on a regularly spaced grid containing $25\%$ of the pixels in the image, and all pixels are used for testing.
We use two metrics, PSNR and \FLIP~\cite{Andersson2020}, for comparisons. \FLIP provides an error map for visualisation and a global measure, i.e., the weighted median of the errors of all pixels. As listed in Table~\ref{table:result}, our method outperforms all compared methods in all datasets and metrics with significant gains. 
In particular, for ``text'' images containing large homogenuous areas and small-sized text, the PSNR of all compared methods ranges from 30.0 to 32.0, while our method is larger than 33.0.

The visual results in Figure~\ref{fig:2dresults} show that all methods can produce realistic images. However, when zoomed in closer, we see that SAPE produces many smoothed patches, while FNN and SIREN produce many random noisy artifacts and SPE contains stripe noise.
Our method reconstructs small-scale structures well without producing much noise.

\begin{table}[t!]
\footnotesize
\tabcolsep=0.07cm
\begin{center}
\begin{tabular}{l>{\columncolor{mygray}}ccc>{\columncolor{mygray}}cc}
\toprule
\multicolumn{1}{l}{\multirow{2}{*}{\centering Model}} & \multicolumn{2}{c}{PSNR$\uparrow$} && \multicolumn{2}{c}{\FLIP$\downarrow$} \\ \cmidrule{2-3} \cmidrule{5-6}
\multicolumn{1}{c}{}  & \multicolumn{1}{>{\columncolor{white}}c}{Natural} & Text && \multicolumn{1}{>{\columncolor{white}}c}{Natural}& Text \\ \midrule
FFN   & 25.57 $\pm$ 4.19   & 30.47 $\pm$ 2.11                     && 0.131$\pm$0.041          & 0.096$\pm$0.043       \\
SIREN & 27.03 $\pm$ 4.28   & 30.81 $\pm$ 1.72                     && 0.114$\pm$0.040          & 0.070$\pm$0.020       \\
SPE   & 26.49 $\pm$ 3.89   & 31.12 $\pm$ 2.18                     && 0.130$\pm$0.038          & 0.065$\pm$0.022       \\
SAPE  & 28.09 $\pm$ 4.04   & 31.84 $\pm$ 2.15 && 0.118$\pm$0.026          & 0.083$\pm$0.041       \\\midrule
FINN  & \textbf{28.51 $\pm$ 4.35}   & \textbf{33.09 $\pm$ 1.97}                     && \textbf{0.100$\pm$0.037}          & \textbf{0.042$\pm$0.016}     \\\bottomrule
\end{tabular}
\caption{Quantitative comparison results for image representation.}
\label{table:result}
\end{center}
\end{table}

\paragraph{Design Choices.}
We validate each component of our filter and show the numerical results of the variants in Table~\ref{table:qr2}. 

\begin{table}[t!]
\footnotesize
\tabcolsep=0.08cm
\begin{center}
\begin{tabular}{l>{\columncolor{mygray}}cc>{\columncolor{mygray}}cc}\toprule
\rowcolor{white}
& FINN$_{\text{so}}$ &FINN$_{\text{ro}}$ &FINN$_{\text{in}}$ &FINN$_{\text{fxs}}$ \\ \midrule
Natural  & 27.91 $\pm$ 4.21 & {28.09 $\pm$ 4.06} &28.26 $\pm$ 4.35 & \textbf{28.59 $\pm$ 4.39} \\
Text  & 31.19 $\pm$ 1.86 & {31.99 $\pm$ 1.73} &  32.22 $\pm$ 2.58    &\textbf{33.17 $\pm$ 2.01}  \\\bottomrule
\end{tabular}
\caption{Quantitative comparison results for FINN variants.}
\label{table:qr2}
\end{center}
\end{table}

\textbf{Use only smoothing or recovering operator}. 
FINN$_{\text{ro}}$ and FINN$_{\text{so}}$ are networks that integrate the recovering and smoothing operators, respectively, into FFN. They perform worse than FINN, but both outperform FFN. This suggests that either noise reduction or detail enhancement contributes to the improvement in reconstruction. However, each operator has only a single effect, smoothing or sharpening the results, so it is not comparable to apply both.

\textbf{Filtering on inputs instead of features of MLPs}. 
We refer to the method that applies the filter to the input Fourier feature $\gamma(x)$ as FINN$_{\text{in}}$. Although FINN$_{\text{in}}$ can adjust Fourier features and performs better than FFN, its performance is not comparable to FINN because the subsequent MLP layers are unfiltered, making the filter less effective. However, FINN$_{\text{in}}$ outperforms FINN$_{\text{ro}}$ and FINN$_{\text{so}}$, suggesting that our filter is more effective than those that can only smooth or recover, even when the filter is applied only to the inputs.

\textbf{Use layer-wise scaling vectors $F_x^i$ instead of a global $F_x$}. 
FINN$_{\text{fxs}}$ means that $F_x$ passed to the different MLP layers are generated independently, i.e., denoted as $F_x^i$ for layer $i$. This adds many more network parameters, but the performance gain over using a global $F_x$ is negligible. The main reason is that $F_x$ and all $F_x^i$ are derived directly from the same Fourier features, which limits their ability to generate diverse scaling vectors. While it is possible to generate more complex $F_x^i$ by customising $\gamma_s(x)$ and $M$, this is not recommended because a stronger recovering operator reduces the smoothing effects. It is also shown that simply increasing the network capacity does not improve the performance.

\begin{figure}[tp!]
\begin{center}
{\includegraphics[width=1.0\linewidth]{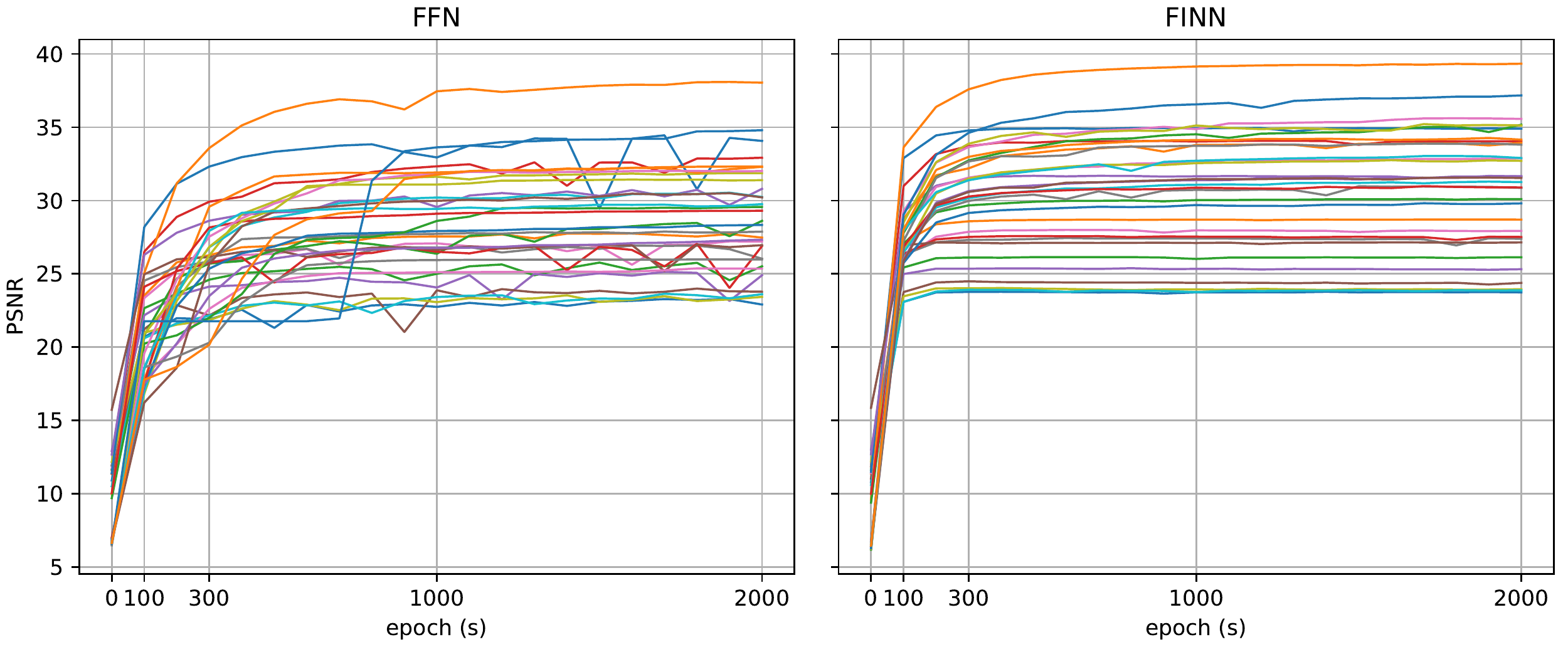} }
\end{center}
\caption{
Convergence and stability comparison with the networks with and without filters. With filter, FINN (right) converges faster, is more stable, and has higher PSNR values than FFN (left).
}
\label{fig:convergence}
\end{figure}

\paragraph{Network Convergence and Stability.}
Figure~\ref{fig:convergence} shows the PSNR curves of 32 images from the ``natural" and ``text" datasets. Both FFN (left) and FINN (right) are trained for 2000 iterations. FINN converges after 300 iterations and maintains stable PSNR values for most images since then. In contrast, FFN converges more slowly and less stably during optimization, resulting in slightly curved and non-monotonic PSNR curves. This suggests that the filter prevents overfitting, leading to better generalization.

\subsection{3D Shape Reconstruction}

For 3D surface reconstruction, we consider the 3D surface as a zero level-set of the signed distance field (SDF), where each 3D query point has a corresponding value indicating its distance and whether it is inside or outside the surface. We train the network to compute the SDF from the input point clouds and their normals. In addition, we apply a regularization term to constrain the gradient of the SDF at all spatial positions to a unit vector, as suggested by~\cite{AtzmonL20,GroppYHAL20}. The loss function is defined as
\begin{equation}
\begin{aligned}
\mathcal{L}_{\text{sdf}} &=\sum_{x\in \Omega}|\|\nabla f_{\theta}(x)\| -1| 
+ \sum_{x\in \Omega\setminus \Omega_0} {\exp(-|f_{\theta}(x)|)} \\
&+ \sum_{x\in \Omega_0}(|f_{\theta}(x)|+\|\nabla f_{\theta}(x) - n(x)\|)
\end{aligned}
\label{eq4}
\end{equation}
where $\Omega$ is the set of points and $\Omega_0$ contains only points on the surface. $f_{\theta}(x)$ and $\nabla f(x)$ are the fitted SDFs and gradients, and $n(x)$ is the ground truth normal of the points. Each loss term is weighted by a constant.

The network for this task is similar to that for image representation, except that the output layer has no activation and the number of hidden layers is increased from 3 to 4.
We use a low frequency, $\sigma=1$ and $\gamma_s(x)=\gamma(x)$, because SDFs are scalar fields, usually much less complex than the RGB color field.
We also reduce the dimension of the Fourier features from 512 to 256. To train the network, in addition to points on the surface, points outside the surface are needed, which are randomly sampled from a bounding cube. The network is trained for 10k iterations using the Adam optimizer with a learning rate of 5$\times$10$^{-4}$. After training, a set of regular grid points is sampled to obtain the SDFs, and the marching cubes algorithm~\cite{lorensen1987marching} is used to extract the triangular mesh.

\begin{figure*}[t!]
\begin{center}
{\includegraphics[width=0.7\linewidth]{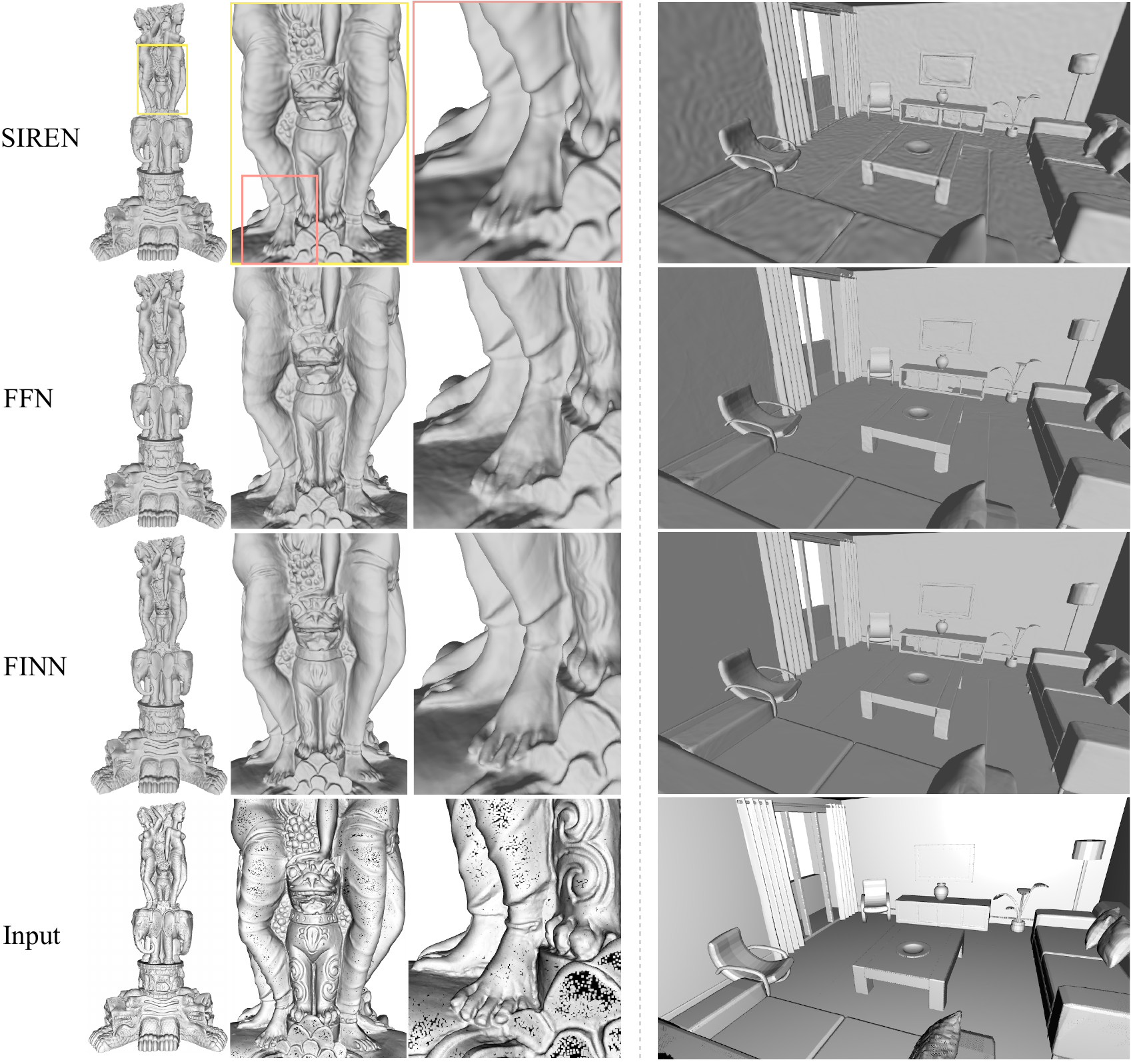} }
\end{center}
\caption{
Qualitative results of large-scale 3D shape reconstruction. For the ``statue" (left) and ``interior room" (right) scenes, FFN and SIREN contain both random high frequencies and excessively smoothed regions in the reconstructed shapes. For example, SIREN has many small bumps in the floor and wall regions of ``interior room", and FFN also contains noisy artifacts in there. Similar problems exist in ``statue", where SIREN and FFN exhibit unexpected noise and lose some sharp geometric features, such as the carved details on the feet that are highlighted. In both scenes, FINN keeps the flat regions flat and restores more sharp features.
}\label{fig:3d}
\end{figure*}

We compare our method with FFN and SIREN on large scenes with several million triangles. Figure~\ref{fig:3d} shows two examples, ``statue'' and ``interior room'', both of which have very complex geometric features and large smooth surface patches. All methods can produce plausible results with fine-grained geometric detail. However, when zoomed in closer, we see that FNN and SIREN produce random bumps and noisy artifacts along the surface and miss some important features. FINN performs better on both flat and sharp surface regions.

\begin{figure*}[h]
\begin{center}
{\includegraphics[width=1.0\linewidth]{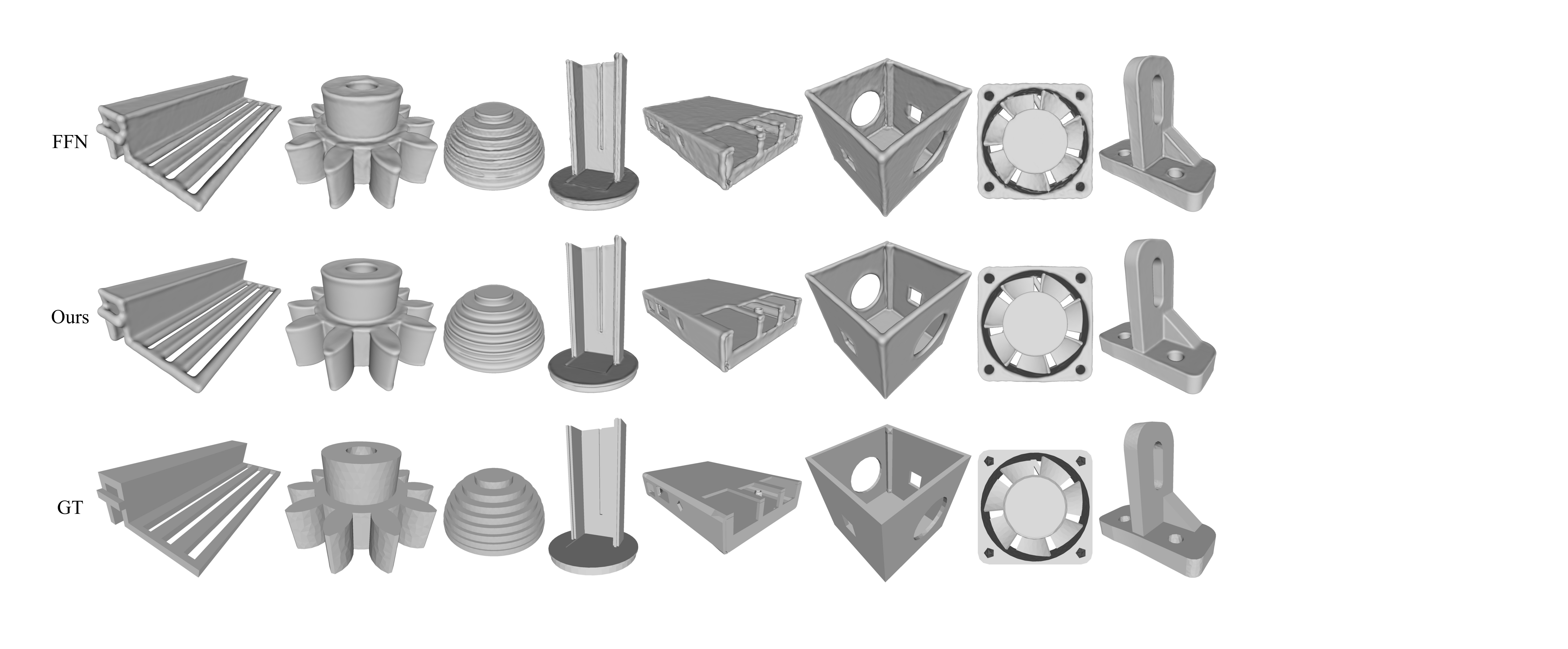} }
\end{center}
\caption{
{Qualitative results from FFN and Ours on the ABC-Dataset. The models contain large flat regions and sharp edges. Ours is better at preserving both flatness and sharpness than FFN in most examples.}
}\label{fig:abc}
\end{figure*}

In addition, we show results on a large scale 3D shape dataset. The ABC dataset~\cite{Koch_2019_CVPR} contains a subset with ground truth surface normals. From this subset, 100 shapes were randomly selected for evaluation. Each shape contains 2048 vertices and was trained and tested using the default setting of our 3D reconstruction neural network. Figure~\ref{fig:abc} shows the results of FFN and Ours, with Ours significantly outperforming FFN. Our method can recover both sharp edges (high frequencies) and flat regions (low frequencies) with high fidelity. This proves that our filter can reduce unexpected noise while improving details. Numerical results are presented in Table~\ref{table:abc}, which show that our method outperforms the FFN method by a large margin.

\begin{table}[tp!]
\footnotesize
\begin{center}
\begin{tabular}{cc}
\toprule
FFN  & 5.69 \\\midrule
Ours    & \textbf{4.81} \\\bottomrule
\end{tabular}
\caption{{Numerical results of 3D reconstruction on ABC-Dataset. Values are chamfer distances multiplied by $10^5$.}}
\label{table:abc}
\end{center}
\end{table}

\begin{figure*}[t!]
\begin{center}
{\includegraphics[width=0.82\linewidth]{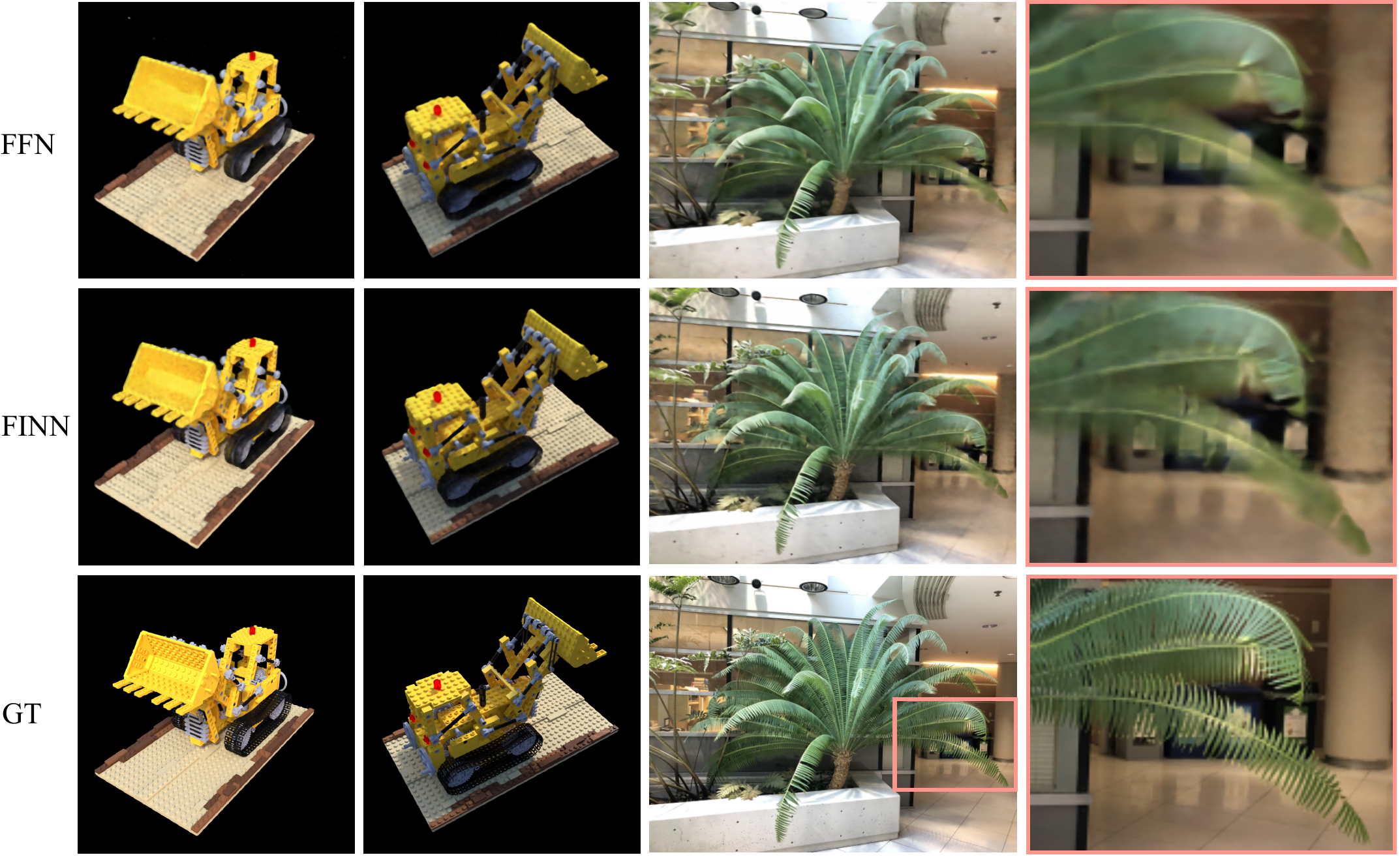} }
\end{center}
\caption{
Qualitative results of novel view synthesis using a ``simplified NeRF''. In the ``Lego'' (left) and ``Fern'' (right) scenes, FINN contributes to a higher level of detail in the results compared to FFN.
}
\label{fig:nerf}
\end{figure*}

\subsection{Novel View Synthesis via 3D Inverse Rendering} 
\label{sec:nerf}
The novel view synthesis task obtains a set of 2D images with known camera poses (i.e., position and view direction) and generates images from new poses. Following Neural Radiance Fields (NeRF)~\cite{MildenhallSTBRN20}, we construct a vector field in 3D space with MLPs that map 3D coordinates to RGB colors and volume densities. Using this field, images with specific poses can be rendered using volume rendering techniques. To train NeRF, we compute the MSE loss between the input images and the re-rendered images from the fitted field. 
Following FFN, we use the ``simplified NeRF" for evaluation, where hierarchical sampling and view dependence have been removed from the original NeRF.

Our network for this task has 512-dimensional Fourier features, MLP with 4 layers, 256 channels, ReLU activation, sigmoid on RGB output, and a scaling vector generator with 512$\times$256 parameters.
Unlike images and 3D shapes, NeRF mixes appearance and geometry information and is therefore more complex. Following images and shapes, which use $\sigma=\sigma_s=10$ for color fields and $\sigma=\sigma_s=1$ for signed distance fields, we set $\sigma\approx10, \sigma_s\approx1$ to optimize neural fields in a wide range, between $\sigma$ and $\sigma_s$.
A better learning scheme would be to decouple appearance and geometry and learn each field in its own frequency spectrum bandwidth. However, this requires a redesign of the framework and we leave this to future research.

We validate our filter on two datasets, including a synthesized scene ``Lego'' and a real scene ``Fern''. The number of images for training and testing are 100, 25 for Lego and 17, 3 for Fern respectively.
FFN sets $\sigma$ by grid search to $6.05$ for Lego and $5.0$ for Fern.
And we set $\sigma,\sigma_s$ to $12,1$. 
Note that we use a 6-layer MLP for FFN, so it has the same network size as FINN. The two networks are trained for 50k iterations using the Adam optimizer with a learning rate of 5$\times$10$^{-4}$.

The results in Table~\ref{table:nerf} show that FINN performs better than FFN for both training and testing for the two datasets, especially for the real scene Fern, with a significant performance advantage. The visual results are shown in Figure~\ref{fig:nerf}, with FINN providing more detail than FFN.
Noise is usually a minor issue in NeRF because each pixel is integrated over a series of samples using volume rendering, averaging local random high frequencies if present.
In practice, using a frequency parameter $\sigma$ that is too high does not lead to random noisy artifacts, but may cause the networks not to converge.

\begin{table}[tp!]
\footnotesize
\tabcolsep=0.07cm
\begin{center}
\begin{tabular}{l>{\columncolor{mygray}}ccc>{\columncolor{mygray}}cc}\toprule
\multicolumn{1}{l}{\multirow{2}{*}{\centering Model}}& \multicolumn{2}{c}{Lego} && \multicolumn{2}{c}{Fern} \\\cmidrule{2-3} \cmidrule{5-6}
\rowcolor{white}
&Train PSNR & Test PSNR &&Train PSNR & Test PSNR \\ \midrule
FFN  & 26.25 $\pm$ 1.00   &25.74 $\pm$ 1.06 && 25.06 $\pm$ 0.44 & 24.30 $\pm$ 0.67   \\
FINN  & \textbf{26.70 $\pm$ 1.00}  & \textbf{26.02 $\pm$ 1.10} && \textbf{25.99 $\pm$ 0.53}  & \textbf{25.28 $\pm$ 0.63} \\\bottomrule
\end{tabular}
\caption{Numerical results of novel view synthesis.}
\label{table:nerf}
\end{center}
\end{table}

\section{Conclusion}

We present a novel filter for neural fields. Our filter gives MLPs better control for smoothing and sharpening neural fields. The filter has two counteractive operators: a smoothing operator that smooths the entire region, and a recovering operator that restores fine details in the oversmoothed regions. Either operator on its own would result in oversmoothing or an increase in noise, while their interaction results in removing much noise while enhancing detail, leading to better fitting and generalization. We demonstrate the effectiveness of our filter on several tasks and show significant improvement over state-of-the-art methods.

\appendix
\setcounter{section}{0}
 \renewcommand{\thesection}{\Alph{section}}

\section{Experimental Details} 
\subsection{Network Details}

\begin{figure*}[h]
\begin{center}
{\includegraphics[width=0.75\linewidth]{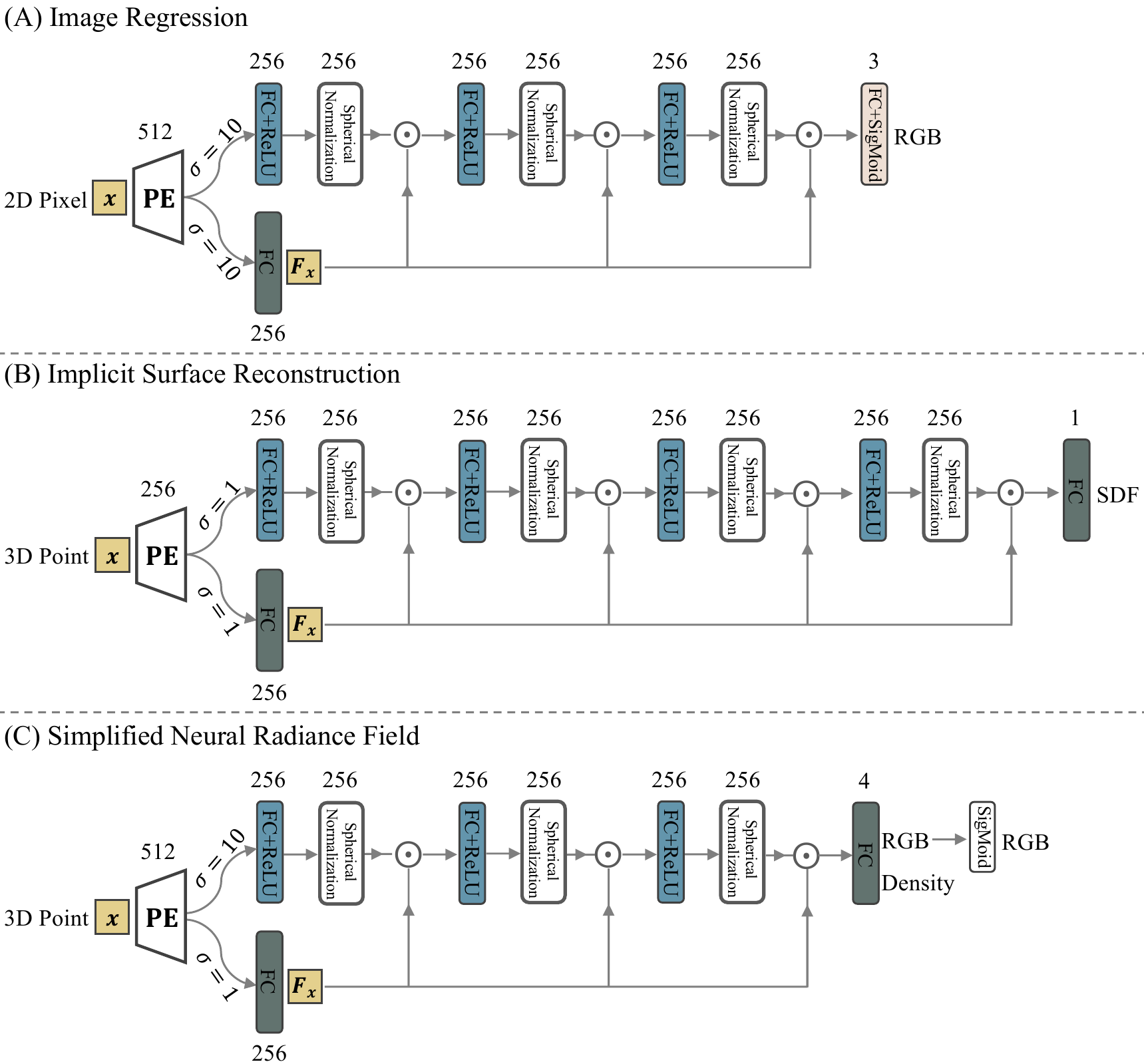} }
\end{center}
\caption{Our networks for 2D image representation, 3D shape reconstruction, and novel view synthesis via 3D inverse rendering.}
\label{fig:nets}
\end{figure*}

Our networks for 2D image representation, 3D shape reconstruction, and novel view synthesis via 3D inverse rendering are detailed in Figure~\ref{fig:nets}.

\paragraph{Impact of Network Size.} 

For most tasks, FFN uses a fixed Fourier feature embedder and 3-layer ReLU MLPs, while SIREN uses 5-layer MLPs with sinusodial activation. The hidden feature size is fixed at 256.
FINN has one more adaptive filter than FFN. Therefore, more network parameters are usually required, with a number of $512\times 256$ and $256\times 256$ for image representation and 3D reconstruction, respectively. 
For the ``simplified NeRF'' that uses 4 MLP layers, we add two additional MLP layers to FFN so that FINN and FFN have the same network size.

\begin{table}[h!]
\footnotesize
\tabcolsep=0.08cm
\begin{center}
\begin{tabular}{lccccc}
\toprule
& FFN &FINN$_{1}$ &FINN$_{2}$ &FINN$_{3}$ &FINN$_{4}$\\ \midrule
Natural  & 25.57 & 28.07&28.43&28.51&28.47  \\
Text  & 30.47  & 32.17&32.91&33.09&33.08   \\\bottomrule
\end{tabular}
\caption{Impact of network size.}
\label{table:t1}
\end{center}
\end{table}

More network parameters can improve performance, but above a certain network size, performance converges. Table~\ref{table:t1} shows the results of FFN and variants of FINN with increasing hidden layers, from 1 to 4. Note that FFN has 3 hidden layers, the same size as FINN$_{1}$, which has only one hidden layer. It is clear that FINN$_{1}$ performs much better than FFN. If the number of hidden layers in FINN is increased from 2 to 4, the results change only slightly.

Our method may require more network parameters, but allows for better generalization that cannot be achieved by simply increasing the size of the network. The methods such as FFN, SIREN, MFN and BACON can already handle data with very small errors or even with errors close to zero. The problem lies in the transformation of the whole space from linear/smooth to strongly nonlinear using too high frequencies. This leads to high generalization error as noisy artifacts increase in the unseen space. Simply increasing the size of the network is not able to improve the generalization. Smoothing regularization is required to reduce noise, but this usually compromises data fitting. Our filter improves generalization without compromising fitting.

\subsection{Additional Training Details}

\paragraph{3D Shape Reconstruction.} 
The batch size when fitting large-scale point clouds is 200k, of which 100k points on the surface are from the input and 100k points off the surface are randomly generated within the unit bounding box.
For medium sized shapes, the number of points for training in a batch is set to 32$^3$.
We generate the signed distance fields with a resolution of 256$^3$ and then extract meshes from them. To calculate the chamfer distance, we uniformly extract 64$^3$ points from the mesh.
The weighting parameters in the loss function are set as follows,
\begin{equation}
\begin{aligned}
\mathcal{L}_{\text{sdf}} &=\sum_{x\in \Omega}a_1|\|\nabla f_{\theta}(x)\| -1|  \\
&+ \sum_{x\in \Omega\setminus \Omega_0}a_2\cdot {\exp(-50|f_{\theta}(x)|)} \notag \\
&+ \sum_{x\in \Omega_0}(a_3|f_{\theta}(x)|+a_4\|\nabla f_{\theta}(x) - n(x)\|)
\end{aligned}
\end{equation}
where $a_1, a_2, a_3$ and $a_4$ are set to $0.1, 0.1, 10$ and $2.0$ respectively. $\sigma$ is set to $1$ for all examples.

\paragraph{NeRF.} 
The ``Lego'' scene datasets contain 100 training images rendered from a synthesis scene with black background colors. ``Fern'' contains 17 training images taken in the real world. The ground truth of the camera poses is computed using Structure from Motion (SfM), which provides the extrinsic parameters of the input images and the 3D positions of the feature points. It is common to use COLMAP~\cite{schoenberger2016sfm} as an open source tool for SfM.
We take 1024 rays with 128 points along each ray in a batch for training. Training takes about an hour for FFN and FINN for 50k epochs.

\section{More Results}

\begin{figure*}[h]
\begin{center}
{\includegraphics[width=0.78\linewidth]{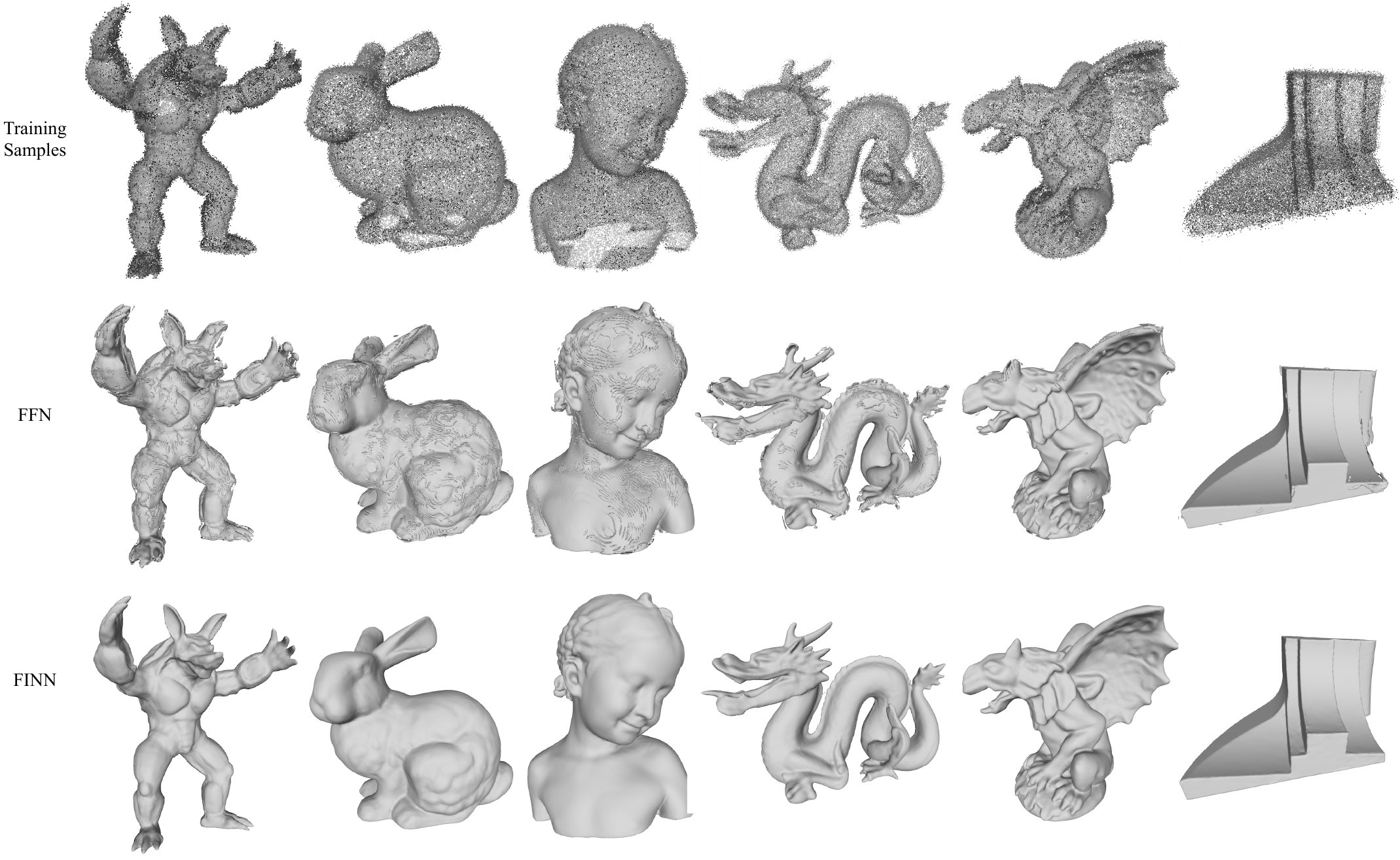} }
\end{center}
\caption{
3D shape reconstruction on noisy point clouds. FFN creates random patches that stick to underlying surfaces while FINN avoids overfitting to the noisy points and create clean shapes.
}
\label{fig:shape2}
\end{figure*}

\paragraph{3D Shape Reconstruction.} 
Additional results on sparse point clouds are shown in Table~\ref{table:qr3}. Each input point cloud contains tens of thousands of points and has some empty holes. FINN generally outperforms FFN for all shapes except ``Bimba'', whose bottom is empty and therefore the reconstruction has large uncertainty. Qualitative results show that both FINN and FFN can fit points and interpolate holes well, but FINN provides more small-scale detail.

\begin{table}[h]
\footnotesize
\begin{center}
\resizebox{\linewidth}{!}{
\begin{tabular}{ccccccc}
\toprule
&Armadillo  &  Bunny & Bimba &Dragon &Gargoyle &Fandisk  \\ \midrule
\multirow{1}{*}{\centering Input} & \multicolumn{6}{c}{\includegraphics[align=c,width=0.9\linewidth]{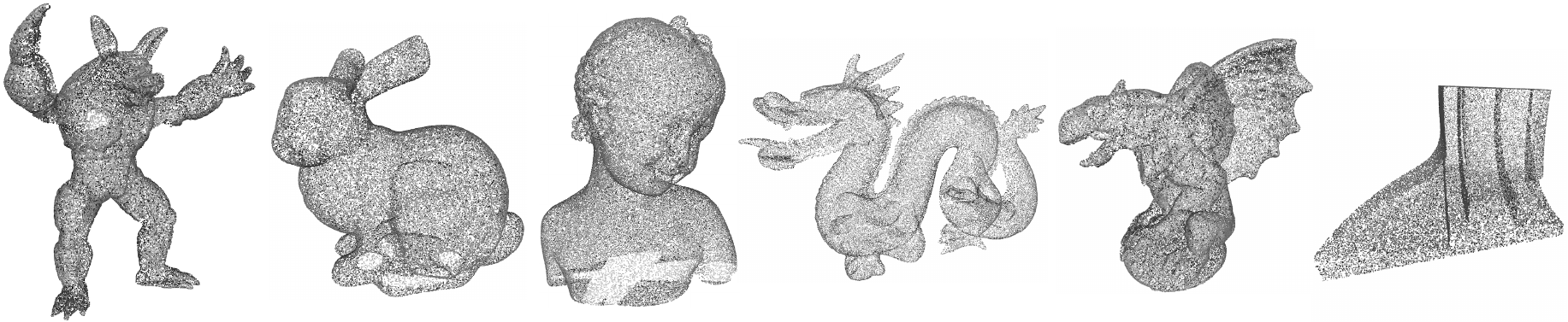}}\\\midrule
\multirow{2}{*}{\centering FFN}& \multicolumn{6}{c}{\includegraphics[align=c,width=0.9\linewidth]{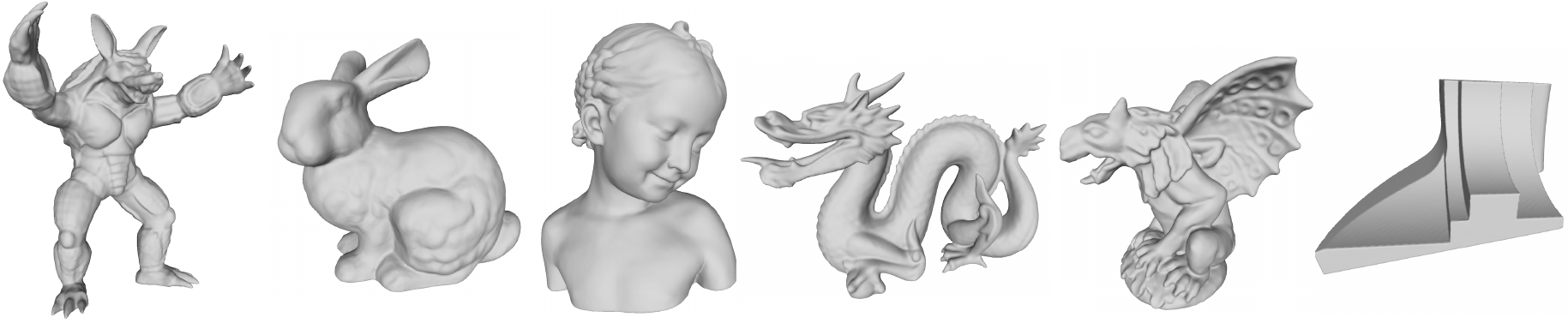}}\\
  & 2.92  &2.04 & \textbf{138.11} & 1.97& 4.58& 1.77 \\\midrule
\multirow{2}{*}{\centering FINN}& \multicolumn{6}{c}{\includegraphics[align=c,width=0.9\linewidth]{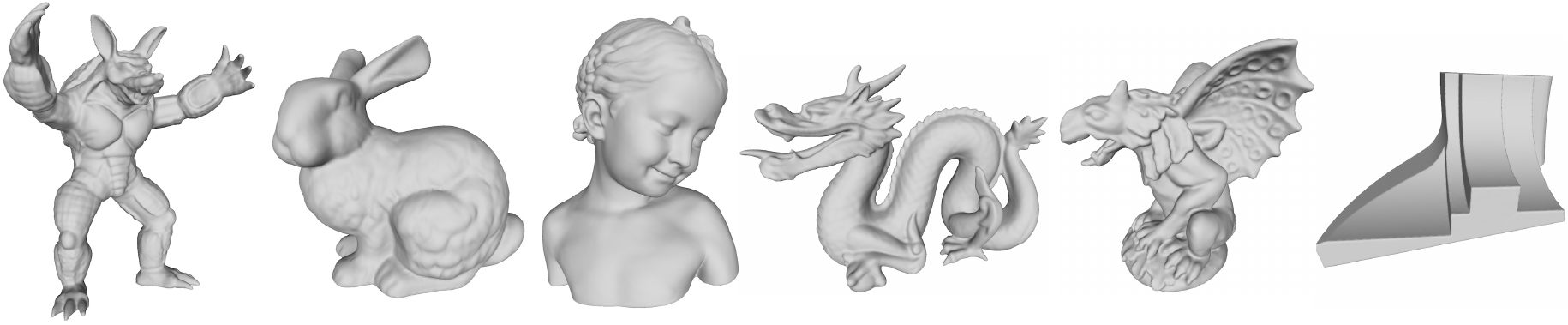}}\\
  &\textbf{2.84}  & \textbf{1.90} & 139.06 & \textbf{1.92} &  \textbf{4.24} & \textbf{1.69}  \\\bottomrule
\end{tabular}
}
\caption{
Results of 3D shape reconstruction of medium sized point clouds. Values are chamfer distances multiplied by $10^5$.}
\label{table:qr3}
\end{center}
\end{table}

\paragraph{Robustness to Noisy Data.}

Random noisy artifacts arise from the use of high-frequency Fourier features in neural fields and are also found in input data; in both cases, our filter can reduce them.
To validate our filter in removing noise in input data, we create noisy point clouds by randomly adding to each point a Gaussian noise of $\mathcal{N}(0,\sigma^2),\sigma=0.02$. Then we take noisy points and orignal points as input and train FFN and FINN with the above experimental settings.
As can be seen in Figure~\ref{fig:shape2}, FINN reconstructs smooth and clean shapes, while FFN overfits noisy points, resulting in noisy patches that adhere to underlying surfaces.
The results show that FINN can handle noisy point sets without the need for explicit regularization constraints.

\section{More Analysis} 

\paragraph{Variants of Smoothing Operators.} 
In addition to the spherical normalization, we add the batch norm for further study. The spherical norm performs better than the batch norm, as shown in Table~\ref{table:t2}. This is because the batch norm has fewer smoothing effects, since it transforms features linearly and is therefore unable to reduce the noise normally caused by dramatic local variations. In contrast, the spherical norm forces features into a hypersphere, which has a strong smoothing effect.

\begin{table}[tp!]
\footnotesize
\tabcolsep=0.1cm
\begin{center}
\begin{tabular}{lcccc}
\toprule
& Batch Norm & Spherical Norm \\ \midrule
Natural / Text  & 27.87 / 31.78 & \textbf{28.51} / \textbf{33.09} \\\bottomrule
\end{tabular}
\caption{Batch Norm VS Spherical Norm.}
\label{table:t2}
\end{center}
\end{table}

\paragraph{Variants of Coordinate Embeddings.} 
We compare two coordinate embeddings, Gaussian Random Fourier Feature (RFF) and Positional Encoding (PosEnc)\cite{tancik2020fourfeat}. We set $\sigma_p$ to 8 and the feature dimension to 512 for PosEnc. Gaussian RFF is actually FFN in this work. Table~\ref{table:t3} shows the results of Gaussian RFF and PosEnc, with and without filters. For both embeddings, the results are significantly improved by applying the filter.

\begin{table}[tp!]
\footnotesize
\tabcolsep=0.06cm
\begin{center}
\begin{tabular}{lcc|cc}
\toprule
& Gaussian RFF & Gaussian RFF+Filter & PosEnc & PosEnc+Filter\\ \midrule
Natural  & 25.57 & \textbf{28.51} & 26.79 & \textbf{27.87} \\
Text  & 30.47 & \textbf{33.09} & 30.53 & \textbf{32.57}\\\bottomrule
\end{tabular}
\caption{Results using different coordinate embedding.}
\label{table:t3}
\end{center}
\end{table}

\paragraph{{Use layer-wise scaling vectors $F_x^i$ instead of a global $F_x$ on 3D reconstruction and novel view synthesis tasks.}}

In addition the image representation, the numerical comparison of using layer-wise recovering operators and a global recovering operator for 3D reconstruction and novel view synthesis is shown in Table~\ref{table:lgcr} and~\ref{table:lgcn}. This shows that adding more scaling parameters does not consistently improve the performance of the different examples. With similar network structure, performance saturates as we increase the number of parameters.

\begin{table}[tp!]
\footnotesize
\begin{center}
\resizebox{\linewidth}{!}{
\begin{tabular}{ccccccc}
\toprule
&Armadillo  &  Bunny & Bimba &Dragon &Gargoyle &Fandisk  \\ \midrule
Global $F_x$  &\textbf{2.84}  & \textbf{1.90} & \textbf{139.06} & \textbf{1.92} &  4.24 & \textbf{1.69}   \\\midrule
Layer-wise $F_x^i$    & 2.90  & 1.99 & 140.07 & 2.03 &  \textbf{4.18} & \textbf{1.69}  \\\bottomrule
\end{tabular}
}
\caption{{Numerical results of 3D reconstruction by using layer-wise recovering operators $F_x^i$ and a global recovering operator $F_x$.}}
\label{table:lgcr}
\end{center}
\end{table}

\begin{table}[tp!]
\footnotesize
\begin{center}
\resizebox{\linewidth}{!}{
\begin{tabular}{l>{\columncolor{mygray}}ccc>{\columncolor{mygray}}cc}\toprule
\multicolumn{1}{l}{\multirow{2}{*}{\centering Model}}& \multicolumn{2}{c}{Lego} && \multicolumn{2}{c}{Fern} \\\cmidrule{2-3} \cmidrule{5-6}
\rowcolor{white}
&Train PSNR & Test PSNR &&Train PSNR & Test PSNR \\ \midrule
Global $F_x$   & \textbf{26.70 $\pm$ 1.00}  & \textbf{26.02 $\pm$ 1.10} && 25.99 $\pm$ 0.53  & 25.28 $\pm$ 0.63 \\
Layer-wise $F_x^i$  & 26.77 $\pm$ 1.03   &25.94 $\pm$ 1.32 && \textbf{26.00 $\pm$ 0.54} & \textbf{25.34 $\pm$ 0.65}   \\
\bottomrule
\end{tabular}
}
\caption{{Numerical results of novel view synthesis by using layer-wise recovering operators $F_x^i$ and a global operator $F_x$.}}
\label{table:lgcn}
\end{center}
\end{table}

\begin{figure}[tp!]
\begin{center}
{\includegraphics[width=1.0\linewidth]{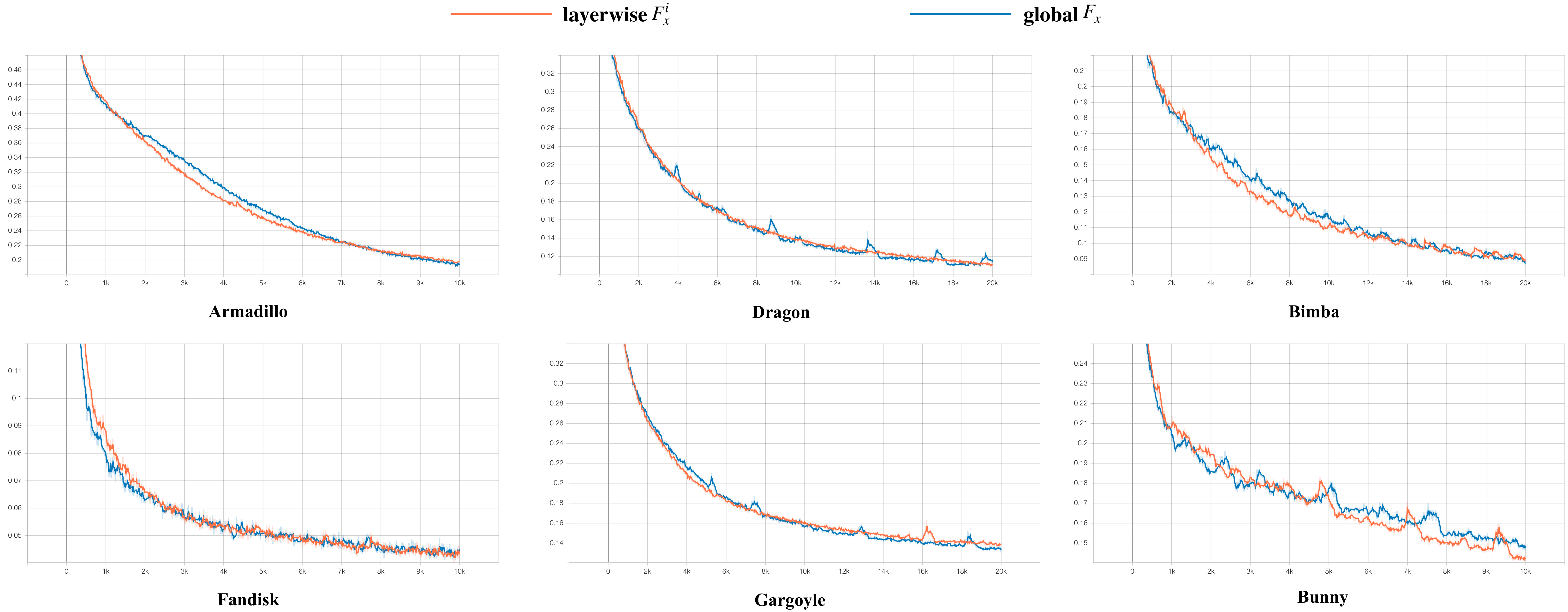} }
\end{center}
\caption{
{Training dynamics with layer-wise $F_x^i$ and global $F_x$ in the 3D reconstruction network. The two variants have similar convergence speed and converge to similar optima.}
}\label{fig:lgc}
\end{figure}

We also show the training dynamics of the 3D reconstruction network. The figure~\ref{fig:lgc} shows that the layer-wise operator and the global operator have quite similar training dynamics and converge to similar optima for six examples.

\begin{figure*}[h]
\begin{center}
{\includegraphics[width=0.75\linewidth]{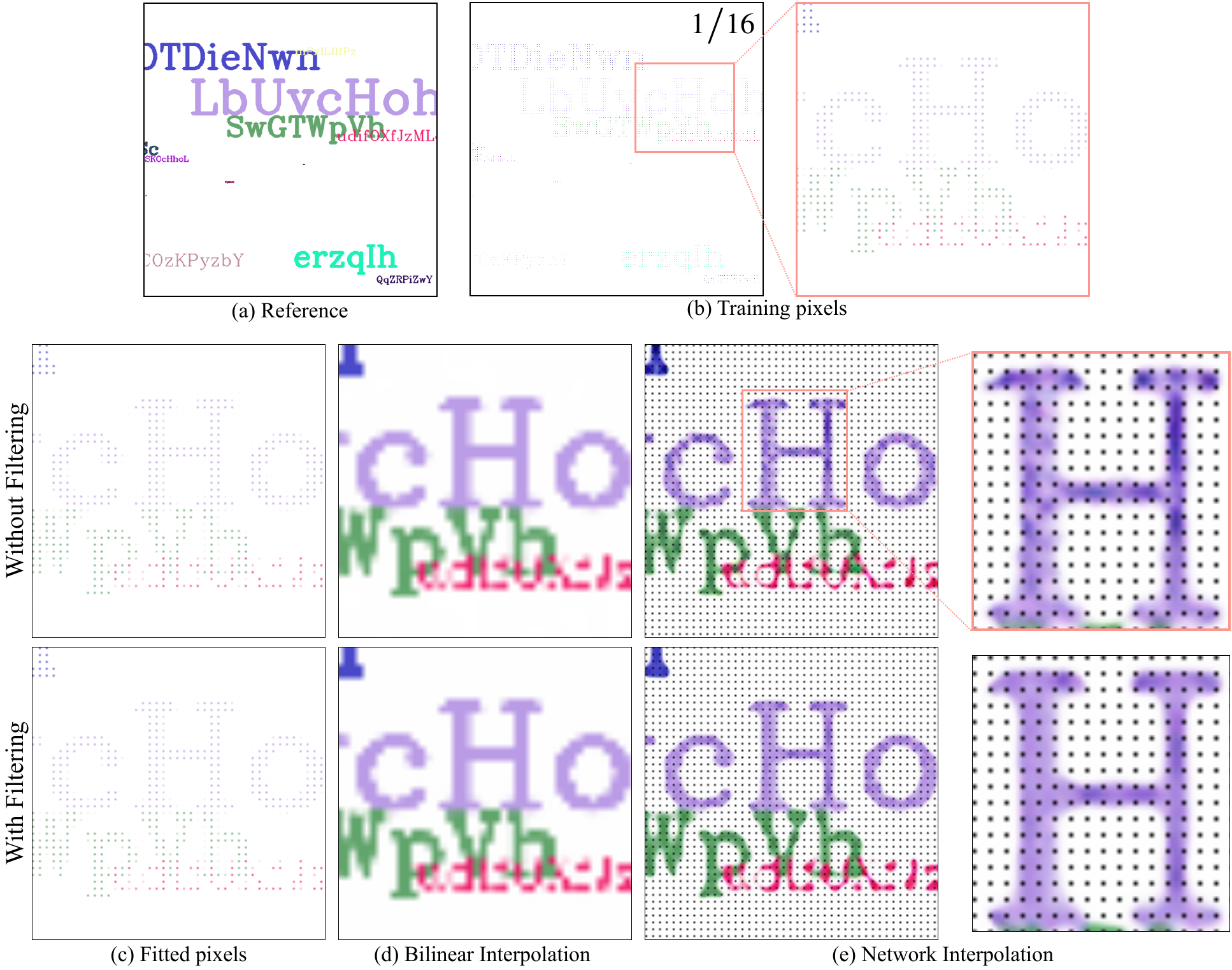} }
\end{center}
\caption{
Illustration of the filter effect. Starting from training pixels in (b) sampled from a reference image (a), we fit (b) with FFN and FINN. We show the results with a subregion of (b) highlighting some texts. The fitted images are shown in (c), which is then extrapolated using bilinear interpolation (see (d)). The results for FFN and FINN in (c)\&(d) are very similar. In contrast, the network interpolants in (e) show very different results. For example, in the highlighted text ``H", FFN randomly deepens or flattens the purple color, while our filter can smooth those noises and produce a clean text. To better see the interpolation, the training pixels in (e) are colored black.
}
\label{fig:interpolant}
\end{figure*}

\paragraph{Filter Effects of Image Interpolation.} 
We fitted an image with a resolution of 128$\times$128 pixels using FINN and FFN. The results in Figure~\ref{fig:interpolant} (c) show that both FFN and FINN can accurately fit the training pixels. We then extrapolate the fit images by a factor of 4 in the vertical and horizontal directions to a resolution of 512$\times$512 pixels. We first use bilinear interpolation, whose results are very similar for the two fitted images, as shown in (d). Compared to the bilinear interpolation, the interpolations with learned interpolants show more local details, as shown in (e). For better illustration of the interpolation, the training pixels are drawn in black. From (e), it can be seen that FFN adds many noisy artifacts by randomly deepening and flattening the colors. This indicates that while FFN allows ReLU-MLP to fit high-frequency training samples, it also introduces high-frequency noise between them. With the filter, FINN reduces variations between nearby training pixels and creates smooth patches within texts, making it more generalizable to unseen pixels.

\bibliographystyle{CVMbib}
\bibliography{refs}

\end{document}